\documentclass[sigconf,nonacm,balance=false]{acmart}
\usepackage{tabularx}
\usepackage{makecell}
\usepackage{enumitem}
\setcopyright{none}
\acmYear{2026}
\acmConference{}{}{}
\acmBooktitle{}
\acmVolume{}
\acmNumber{}
\acmDOI{}
\acmISBN{}
\makeatletter
\renewcommand{\@authorfont}{\large\bfseries}
\renewcommand{\@affiliationfont}{\small\normalfont}
\makeatother
\title{NMIXX: Domain-Adapted Neural Embeddings for Cross-Lingual eXploration of Finance}
\author{Hanwool Lee}
\authornote{These authors contributed equally to this research.}
\affiliation{
  \institution{AIM Intelligence}
  \institution{Seoul National University}
  \country{South Korea}}
\email{gksdnf424@gmail.com}

\author{Sara Yu}
\authornotemark[1]
\affiliation{
  \institution{KT}
  \country{South Korea}}
\email{sara.yu@kt.com}

\author{Yewon Hwang}
\authornotemark[1]
\affiliation{
  \institution{Kakao Mobility}
  \country{South Korea}}
\email{hwyewon@gmail.com}

\author{Jonghyun Choi}
\affiliation{
  \institution{Samsung Fire \& Marine Insurance}
  \country{South Korea}}
\email{excelsiorcjh@gmail.com}

\author{Heejae Ahn}
\affiliation{
  \institution{KB Securities}
  \country{South Korea}}
\email{se21262@kbfg.com}

\author{Sungbum Jung}
\affiliation{
  \institution{Netmarble}
  \country{South Korea}}
\email{successtiger@netmarble.com}

\author{Youngjae Yu}
\authornote{Corresponding author.}
\affiliation{
  \institution{Seoul National University}
  \country{South Korea}}
\email{youngjae4yu@gmail.com}

\renewcommand{\shortauthors}{Lee et al.}

\hypersetup{pdfauthor={Hanwool Lee, Sara Yu, Yewon Hwang, Jonghyun Choi, Heejae Ahn, Sungbum Jung, Youngjae Yu},pdfkeywords={Financial embeddings, Domain adaptation, Korean, Semantic textual similarity},pdfpublisher={The authors},pdfpublication={Preprint},pdfpubtype={report}}
\begin{document}
\raggedbottom
\begin{abstract}
Financial text embeddings must distinguish changes in event status, perspective, and obligations even when passages share similar wording. NMIXX adapts existing encoders through 18.8k source-linked triplets: paraphrases and Korean--English translations preserve meaning, while targeted financial rewrites introduce semantic contrasts. We examine this recipe across seven backbones on English and Korean financial and general-domain semantic textual similarity (STS), and analyze the composition and passage lengths of KorFinSTS. BGE-M3 attains the highest adapted financial correlations in this comparison, improving from 0.1969 to 0.2967 on FinSTS and from 0.0512 to 0.2732 on KorFinSTS. Its general English and Korean correlations decrease by 0.0391 and 0.0463. Across the seven models, five improve their mean financial correlation, but all reduce their mean general-domain correlation. Per-language comparisons and benchmark-weight sensitivity analysis reveal differences obscured by a single aggregate score. The study contributes a finance-specific supervision design and evidence for evaluating adaptation jointly with retained general semantic capability; direct cross-language retrieval remains outside its evaluation scope.
\end{abstract}


\maketitle

\section{Introduction}
Financial texts can be lexically similar yet differ in economically consequential ways: a transaction may be planned rather than completed, an analyst's projection may be mistaken for an observed fact, and a regulatory condition may change an obligation. Sentence embeddings must preserve these distinctions if they are to support financial information access. General semantic similarity performance alone provides limited evidence that an encoder captures such differences in a specialized domain~\citep{tang2025finmtebfinancemassivetext,hwang2025twiceadvantageslowresourcedomainspecific}.

The challenge is particularly relevant to Korean financial text. English-centric evaluation can overlook language-specific terminology and register, while translated evaluation data may not preserve the distinctions found in native financial documents. Financial benchmarks such as FinMTEB and KorFinMTEB motivate domain- and language-aware evaluation~\citep{tang2025finmtebfinancemassivetext,hwang2025twiceadvantageslowresourcedomainspecific}. For an applied financial system, however, a second question is equally important: how much general semantic capability is retained after adaptation?

\begin{figure*}[!t]
\centering
\includegraphics[width=\textwidth]{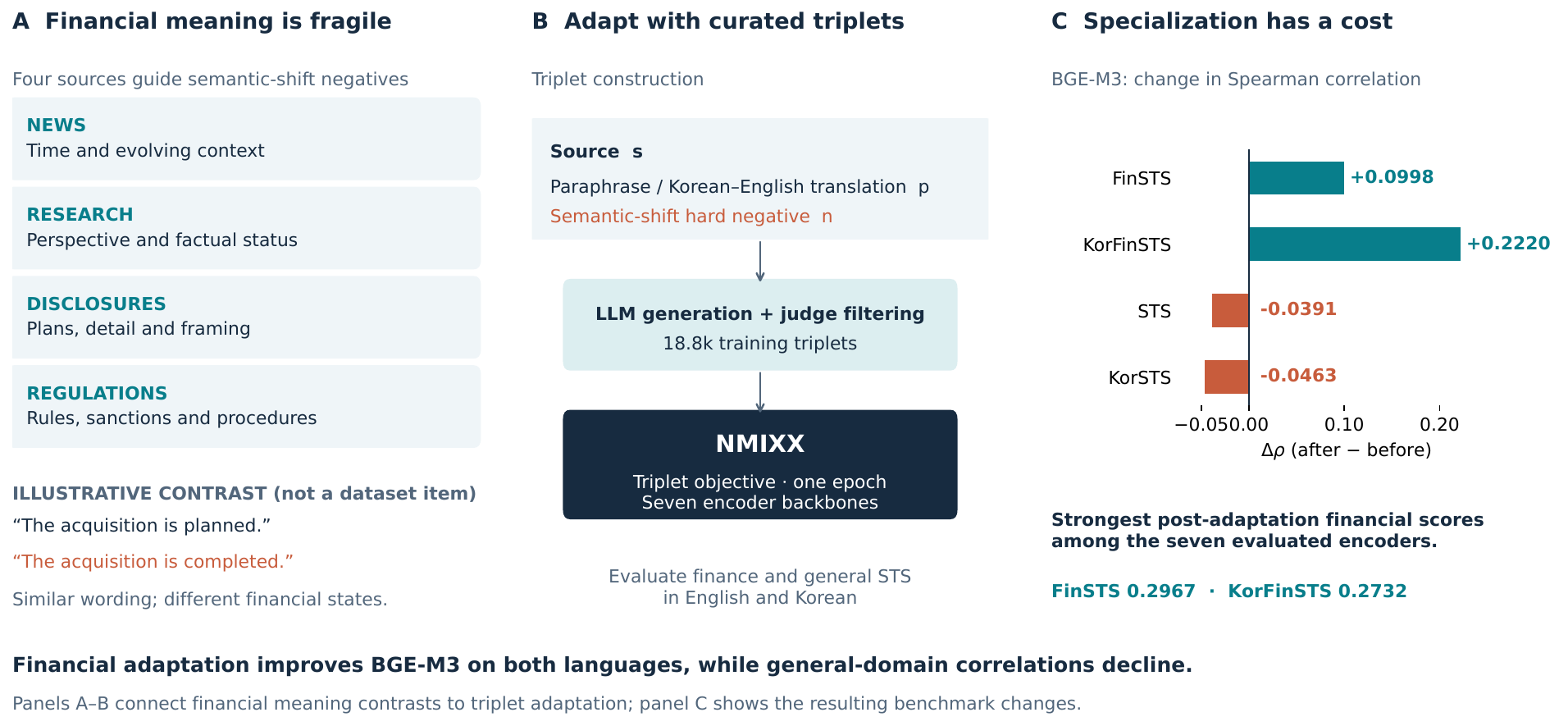}
\caption{NMIXX at a glance. (A) Financial document types motivate semantic-shift negatives; the sentence contrast is illustrative. (B) Triplet construction and adaptation of existing encoders. (C) BGE-M3's changes on all four benchmarks connect the two research questions: financial improvements coexist with general-domain losses.}
\Description{Three panels show financial semantic distinctions, the NMIXX triplet pipeline, and BGE-M3 changes of plus 0.0998, plus 0.2220, minus 0.0391, and minus 0.0463 on the four benchmarks.}
\label{fig:main}
\end{figure*}

We study \textbf{N}eural e\textbf{M}bedd\textbf{I}ngs for \textbf{X}-lingual e\textbf{X}ploration of Finance (NMIXX), a triplet-based framework that specifies which financial meanings an embedding should preserve and which changes it should separate. Each source statement is paired with a meaning-preserving paraphrase or Korean--English translation and a lexically similar, meaning-changing rewrite. The contribution is a supervision design for existing encoders. Figure~\ref{fig:main} connects this construction strategy to the observed specialization trade-off.

Our evaluation addresses two questions: \textbf{RQ1}, does the same adaptation recipe improve financial STS consistently across backbones and languages; and \textbf{RQ2}, how much general-domain STS capability is retained? Seven encoders and four benchmark settings expose these dimensions separately. KorFinSTS adds Korean financial material from news, disclosures, investment research, and legal texts. We examine its source composition and passage-length variation to characterize the dataset alongside model performance.

BGE-M3 achieves the highest adapted correlations on both financial suites, yet loses correlation on both general suites. Other backbones trade improvement in one financial language for regression in the other. Our contributions are: (1) source-linked financial triplet construction; (2) KorFinSTS and a bilingual financial/general comparison across seven encoders; and (3) an analysis of dataset composition and passage lengths, together with per-language gain--retention and benchmark-weight sensitivity analyses. These views connect the supervision design, the data characteristics, and the consequences of adapting a shared encoder.

\section{Related Work}
\paragraph{General and multilingual embeddings.}
Sentence-BERT makes sentence comparison efficient through independent encoding, while SimCSE develops contrastive sentence representation learning~\citep{reimers2019sentencebertsentenceembeddingsusing,gao2022simcsesimplecontrastivelearning}. E5 and GTE extend this approach through weak supervision and multi-stage training~\citep{wang2024textembeddingsweaklysupervisedcontrastive,li2023generaltextembeddingsmultistage}. Multilingual E5 combines large-scale multilingual pair pre-training with labeled fine-tuning~\citep{wang2024multilinguale5}; BGE-M3 jointly supports multilingual, multi-granularity, and multiple retrieval representations~\citep{chen2024bgem3}. These are foundations on which a domain adaptation recipe can operate. NMIXX changes the training contrasts supplied to existing encoders, rather than introducing a new encoder architecture. Its Korean/English setting tests adaptation across two languages; separate within-language STS scores do not establish cross-language retrieval effectiveness.

\begin{figure*}[!t]
\centering
\includegraphics[width=\textwidth]{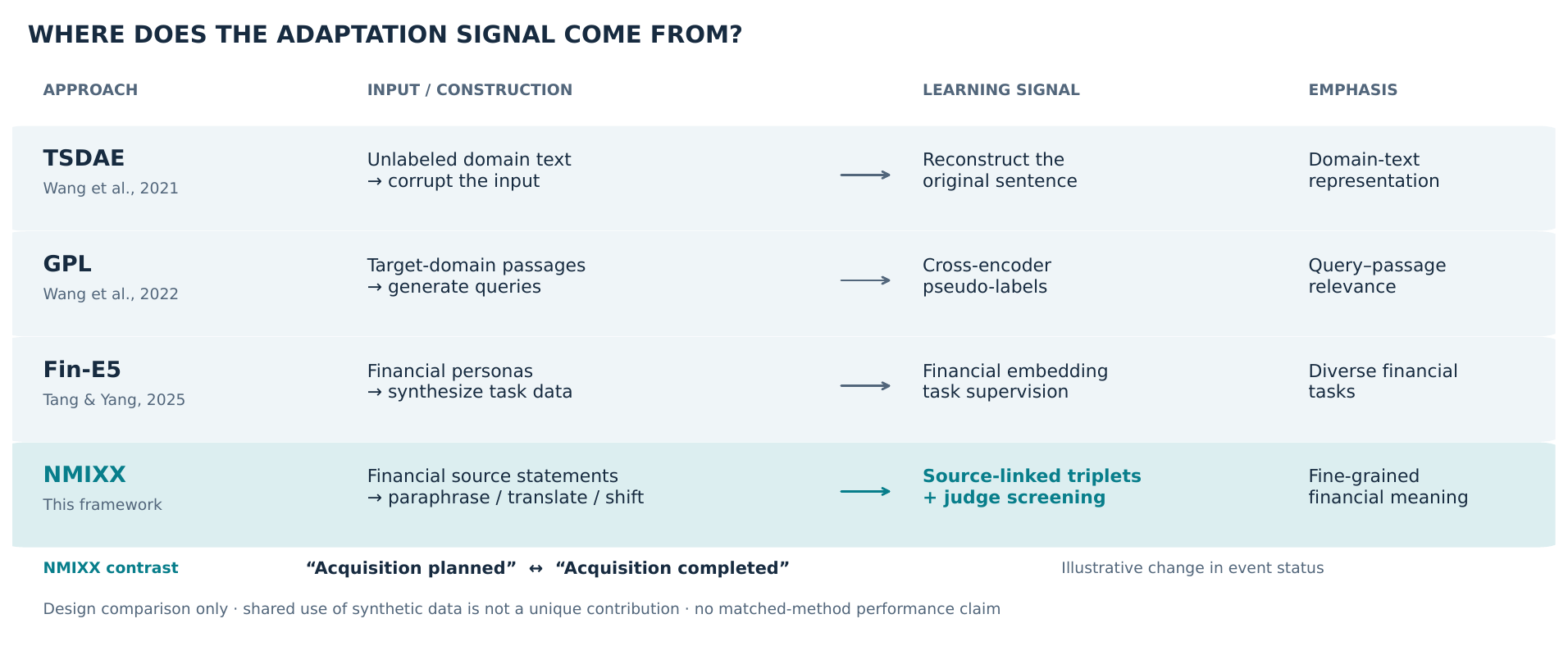}
\caption{Positioning by supervision construction. TSDAE~\citep{wang2021tsdae}, GPL~\citep{wang2022gpl}, and Fin-E5~\citep{tang2025finmtebfinancemassivetext} illustrate distinct adaptation routes. NMIXX emphasizes financial meaning changes within source-linked triplets, alongside paraphrase/translation positives. These are design emphases, not exclusive capabilities or a performance ranking. The sentence contrast is illustrative.}
\Description{Four horizontal routes compare denoising reconstruction in TSDAE, generated queries and teacher labels in GPL, persona-based financial task synthesis in Fin-E5, and source-linked financial semantic-shift triplets in NMIXX.}
\label{fig:positioning}
\end{figure*}

\paragraph{Unsupervised domain adaptation.}
TSDAE learns sentence representations through denoising reconstruction of unlabeled text and evaluates their usefulness for domain adaptation~\citep{wang2021tsdae}. GPL instead constructs retrieval supervision using generated queries and cross-encoder pseudo-labels~\citep{wang2022gpl}. These approaches show that target-domain exposure can be turned into different learning signals. NMIXX uses a source financial statement as the anchor, a paraphrase or translation as a positive, and a meaning-changing rewrite as a negative. Its emphasis is therefore on a specific semantic distinction within a financial narrative, whereas GPL's generated query--passage relations express retrieval relevance. The two objectives need not be interchangeable: two passages may concern the same query while differing in whether an event has occurred. Figure~\ref{fig:positioning} locates this distinction at the supervision-construction stage.

\paragraph{Synthetic supervision and hard negatives.}
Wang et al.~\citep{wang2023improving} demonstrate synthetic embedding-task generation across many languages using LLMs. Synthetic supervision is therefore an established ingredient. NV-Retriever addresses false negatives through positive-aware relevance thresholds during mining~\citep{moreira2024nvretriever}. NMIXX instead generates targeted financial changes and screens them with a judge. Mining asks which existing passage supplies a useful contrast; this construction asks which alteration of a source statement changes its financial interpretation. SDJC also constructs meaning-changing synthetic contrasts for Japanese domain adaptation~\citep{chen2025domainadaptationjapanesesentence}. NMIXX specializes this construction through financial transformation rules and Korean--English positives. Generation proposes the intended change; judge screening checks its execution and the preservation of unrelated content. Judge acceptance remains a construction check rather than calibrated confidence in label validity.

\paragraph{Financial representations and evaluation.}
FLUE motivates financial language evaluation beyond generic benchmarks~\citep{shah2022fluemeetsflangbenchmarks}. FinMTEB spans financial embedding tasks and introduces Fin-E5 with persona-based synthetic training~\citep{tang2025finmtebfinancemassivetext}. Fin-E5 is a close domain-adaptation reference: it emphasizes diverse financial tasks, while NMIXX emphasizes financial semantic-shift contrasts and Korean/English STS. Liu et al.~\citep{lee2024financialsts} directly study subtle changes in financial narratives, providing the closest task-level motivation. NMIXX extends this concern to native Korean financial sources and a shared adaptation recipe across seven backbones. TWICE further motivates evaluation in low-resource domain settings~\citep{hwang2025twiceadvantageslowresourcedomainspecific}. These comparisons position the supervision design; our experiments compare backbones rather than adaptation methods under matched data and compute.

\paragraph{Scope of the present comparison.}
MTEB and MMTEB encourage evaluation across tasks and languages~\citep{muennighoff2023mtebmassivetextembedding,enevoldsen2025mmtebmassivemultilingualtext}. Here, the finance/general distinction tests specialization, while the English/Korean distinction tests whether its observed benefits extend across the two evaluated languages. Reading both dimensions together distinguishes financial gains from general semantic retention. Tokenization remains one possible influence among several confounded backbone differences~\citep{wu2020languagescreatedequalmultilingual,wegmann2025tokenizationsensitivelanguagevariation}. The comparison consequently examines the response of existing backbones to a shared supervision recipe.

\section{NMIXX: Financial Semantic Adaptation}
\subsection{Data Construction}
\label{sec:data}
\begin{figure*}[!t]
\centering
\includegraphics[width=\textwidth]{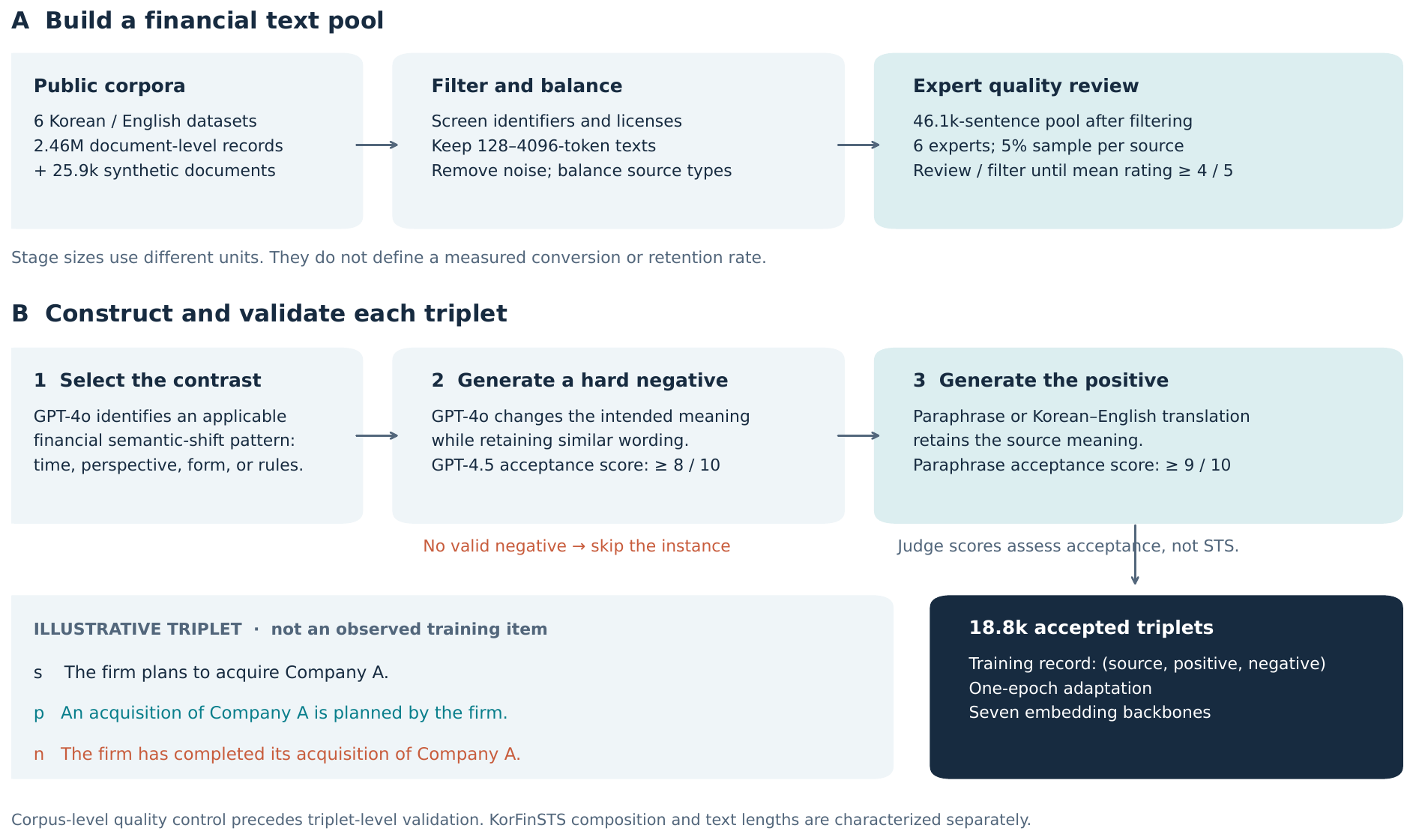}
\caption{NMIXX training-data construction. (A) Public corpora and synthetic augmentation pass through filtering, balancing, and expert review. (B) Semantic-shift negatives and meaning-preserving positives are generated and screened to form triplets. Stage counts use different units and do not imply a conversion rate. The sentence triplet illustrates the intended relation between anchor, positive, and negative.}
\Description{A two-stage pipeline shows 2.46 million raw records plus 25.9 thousand synthetic documents, a filtered pool of 46.1 thousand sentences, expert review, semantic-shift negative and paraphrase generation, judge thresholds of 8 and 9, and 18.8 thousand training triplets.}
\label{fig:data}
\end{figure*}

Figure~\ref{fig:data} expands the construction pipeline, separating corpus-level quality control from instance-level triplet validation. The two stages address different failure modes: source filtering limits unsuitable text, while semantic-shift generation supplies contrasts that retain lexical similarity but alter financial meaning.

\paragraph{Collecting target domain corpora.}
We aggregate six public corpora from Hugging Face---three Korean and three English---totaling 2.46M document--level records prior to filtering (Table~\ref{tab:raw}).

\begin{table}[!ht]
\caption{Public corpora collected before filtering.}
\label{tab:raw}
\small
\setlength{\tabcolsep}{3pt}
\begin{tabularx}{\columnwidth}{@{}X l r l@{}}
\toprule
\textbf{Dataset} & \textbf{Lang.} & \textbf{Rows (k)} & \textbf{License}\\
\midrule
sujet-finance-instruct & ko & 178  & Apache-2.0 \\
finance-legal-mrc    & ko & 359  & CC-BY-SA 4.0   \\
KorFin-ASC           & ko & 8.8  & Apache-2.0         \\
finance-embeddings-\allowbreak investopedia& en & 206  & CC-BY-NC 4.0 \\
fingpt-sentiment-train & en & 76.8 & MIT   \\
FNSPID               & en & 1,630& CC-BY-NC 4.0   \\
\bottomrule
\end{tabularx}
\end{table}

\paragraph{Domain-balanced augmentation.}
Public data are news-heavy; disclosures, sector reports, and broker research are under-represented. We therefore scraped Korean and U.S. regulatory filings and commissioned GPT-4o rewrites to create 25.9k synthetic documents that mirror formal disclosure style while preserving factual content. Samples failing manual spot checks (e.g., hallucinated numbers) were discarded.

\paragraph{Legal and ethical filtering.}
The pipeline excludes personal identifiers and screens source licenses. Table~\ref{tab:raw} includes noncommercial and share-alike terms; public accessibility does not imply unrestricted redistribution. Retained records remain subject to source-specific permissions and attribution requirements.

\paragraph{Pre-processing and quality control.}
We tokenize with the \textsc{Qwen2.5} tokenizer and drop (i) texts shorter than 128 tokens or longer than 4096 tokens and (ii) samples containing HTML artifacts, OCR noise, or high typo entropy. After filtering and per-class down-sampling to match the smallest group (regulatory disclosures), the data pool comprises 46.1k sentences.

\paragraph{Human validation.}
Six domain experts assessed corpus quality through sampled review. Each held at least a Master's degree in economics or business administration or worked in the financial sector (securities, banking, or insurance). For each source, 5\% of the corpus was randomly sampled and rated on a 5-point scale for (1) factual correctness, (2) practical relevance to financial operations, and (3) overall data quality. Sources were iteratively filtered and re-evaluated until the average score across all criteria reached at least 4. This review assesses source material before triplet generation.

\paragraph{Triplet mining for supervised contrastive learning.}
We couple \emph{LLM-as-judge}~\citep{kim2024prometheusinducingfinegrainedevaluation} filtering with a multi-step generation pipeline:

\begin{enumerate}
\item \textbf{Axis identification}: GPT-4o tags each source sentence with the semantic shift pattern(s) it can express meaningfully.
\item \textbf{Hard-negative generation}: Guided by the chosen pattern, GPT-4o synthesizes a semantically divergent yet lexically similar variant; if such a variant cannot be crafted for that pattern, the instance is skipped.
\item \textbf{Hard-negative validation}: GPT-4.5 (LLM-as-judge) scores each \emph{(source, negative)} pair on a 0--10 scale; only pairs with scores $\ge 8$ are retained.
\item \textbf{Positive generation \& validation}: For every accepted source sentence, GPT-4o produces a semantically equivalent paraphrase. GPT-4.5 (LLM-as-judge) keeps only \emph{(source, positive)} pairs scoring $\ge 9$.
\end{enumerate}

This pipeline yields \textbf{18.8k} judge-screened triplets (source, positive, hard negative). Positives also include Korean--English translations, which express the source meaning in the other language. The comparison evaluates the combined positive set rather than the contribution of each positive type separately.

\paragraph{Generation and screening criteria.}
The generation prompts specify a semantic transformation while preserving style and unrelated numerical or accounting facts. The judge checks the requested modification, preservation of other content, and stylistic consistency. A targeted hypothetical change, such as changing a planned event into a completed event, defines a contrastive example rather than a factual claim about the company. Judge scores assess construction quality, whereas the STS benchmarks evaluate the resulting representations. Keeping these roles distinct avoids treating an accepted rewrite as a measured improvement in embedding quality.

\subsection{Financial Semantic-Shift Typology}
Building on financial narrative STS~\citep{lee2024financialsts}, we organize transformations by four source types. The operative distinction is between changing expression and changing the represented financial proposition. A positive should preserve the event, attribution, and obligations; a negative should alter a specified interpretation while retaining unrelated information. The source types organize candidate transformations rather than mutually exclusive semantic labels. Here ``shift'' denotes a constructed meaning contrast, not longitudinal language drift.

\paragraph{Financial news: temporal variation.}
News motivates contrasts in event timing and evolving context, such as whether a transaction is expected or has been completed. Such a change affects the represented event state. A stylistic reformulation that preserves that state should remain a positive rather than become a negative merely because its wording differs.

\paragraph{Investment research: perspectival framing.}
Research reports motivate contrasts between company-level and macroeconomic analysis, reported facts and analyst opinions, and financial versus everyday senses of terms such as ``consensus.'' These distinctions guide generation; the experiment does not report effectiveness per pattern.

\paragraph{Disclosures: structural and contextual meaning.}
Four patterns target intensified sentiment, elaborated details, plan realization, and emerging situations. Examples include significant versus modest growth, or a planned transaction becoming canceled. The transformation should change the specified interpretation while retaining unrelated source information.

\paragraph{Legal texts: rules and obligations.}
Legal sources motivate changes in interpretation, sanction application, and procedural requirements. These contrasts can alter obligations despite similar wording. They are training constructions, not validated legal interpretations, and their individual contribution is not isolated by the experiment.

\section{KorFinSTS}
We construct KorFinSTS for Korean financial semantic textual similarity using four sources: financial news, regulatory disclosures, investment research, and legal or regulatory texts. Construction follows the corpus sourcing process in Section~\ref{sec:data}, followed by expert review of sentence pairs. Native Korean examples target register and terminology that may be obscured by direct translation~\citep{hwang2025twiceadvantageslowresourcedomainspecific}.

\begin{figure}[!ht]
\centering
\includegraphics[width=\columnwidth]{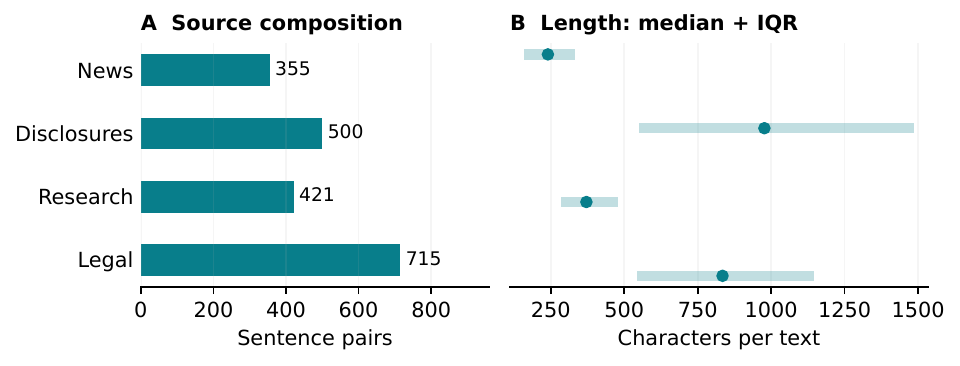}
\caption{KorFinSTS dataset analysis. (A) Pair counts by source. (B) Median and interquartile range of text length, pooling both sentence columns within each source. Characters include stored whitespace; lengths are not model-token counts.}
\Description{Counts are 355 news, 500 disclosure, 421 research, and 715 legal pairs. Median character lengths are 240, 977, 371, and 834.5 respectively.}
\label{fig:korfin-profile}
\end{figure}

\paragraph{Dataset composition.}
We analyze KorFinSTS to characterize its source composition and passage lengths. The dataset snapshot used for this profile contains 1,991 pairs: 715 legal pairs (35.9\%), 500 disclosure pairs (25.1\%), 421 investment-research pairs (21.1\%), and 355 news pairs (17.8\%). Its score field is binary, with 822 zeros and 1,169 ones; the mean and sample standard deviation are 0.587 and 0.492. Figure~\ref{fig:korfin-profile} summarizes the source distribution and text-length variation.

\paragraph{Length and financial document characteristics.}
Figure~\ref{fig:korfin-profile} shows marked length differences. Counting Unicode characters in the stored text, including whitespace, median lengths are 240 for news, 371 for investment research, 977 for disclosures, and 834.5 for legal texts. Both sentence columns are pooled without deduplicating repeated text occurrences. Disclosure texts are thus roughly four times as long as news at the median, making this benchmark a mixture of short and extended financial passages despite its STS framing.

The source categories also organize different interpretation requirements: news supplies event-centered narratives; investment research combines performance discussion and expectations; disclosures emphasize company-specific reporting; and legal texts express conditions and obligations. These descriptions motivate the source-aware construction in Section~3 rather than constituting newly annotated semantic categories. The profile identifies two dimensions of the dataset: financial register and passage length. It characterizes the financial text collection without establishing per-source performance or a measured truncation effect.

\section{Experiments}

\subsection{Experimental Design}
We adapt seven encoders on 18.8k triplets $\langle s,p,n\rangle$. Positives preserve meaning through an in-domain paraphrase or Korean--English translation; negatives apply the semantic-shift rules in Section~3. Translation positives encourage bilingual alignment, while the evaluation measures the resulting representations on separate within-language tasks.

We adopt a temperature-scaled triplet negative-log-likelihood loss:
\begin{equation}
\mathcal{L} = -\log\frac{\exp(\cos(h_s, h_p)/\tau)}{\exp(\cos(h_s, h_p)/\tau) + \exp(\cos(h_s, h_n)/\tau)},
\end{equation}
where $h_s$, $h_p$, and $h_n$ denote encoder outputs and $\tau>0$ controls the scale. The loss decreases as the positive--negative cosine-similarity margin increases. Source-linked contrasts thus specify the meaning distinction the objective is asked to learn; translation positives apply meaning preservation across Korean and English. This explains the supervision target without attributing the observed gains to any single component.

\paragraph{Baseline Models.}
We select the seven public checkpoints in Table~\ref{tab:baselines} with reference to MTEB, FinMTEB, and KorFinMTEB~\citep{muennighoff2023mtebmassivetextembedding,tang2025finmtebfinancemassivetext,hwang2025twiceadvantageslowresourcedomainspecific}. Each is adapted with the same triplet objective using four A100 GPUs. Their different architectures, pre-training, and language coverage allow us to examine how the same supervision recipe affects heterogeneous backbones.

\begin{table}[!ht]
  \centering
  \caption{Baseline embedding models, licenses, and language support.}
  \label{tab:baselines}
  \small
  \begin{tabularx}{\columnwidth}{@{}X l l@{}}
    \toprule
    \textbf{Model} & \textbf{License}  & \textbf{Language Support} \\
    \midrule
    bge-en-icl & Apache-2.0 & Mainly English \\
    gte-Qwen2-1.5B-instruct & Apache-2.0 & English \& Chinese \\
    e5-mistral-7b-instruct & MIT & Mainly English \\
    bge-large-en-v1.5 & MIT & Mainly English \\
    all-MiniLM-L12-v2 & Apache-2.0 & Mainly English \\
    instructor-base & Apache-2.0 & Mainly English \\
    bge-m3 & MIT & Multilingual \\
    \bottomrule
  \end{tabularx}
\end{table}

\begin{table*}[!t]
  \centering
  \caption{Spearman's $\rho$ on four STS benchmarks before and after domain adaptation. The higher value of each before/after pair is \textbf{bolded}. Boldface denotes a numerical comparison, not statistical significance.}
  \label{tab:exp_results}
  \begin{tabular}{@{}l cc cc cc cc@{}}
    \toprule
     & \multicolumn{2}{c}{\textbf{FinSTS}}
     & \multicolumn{2}{c}{\textbf{KorFinSTS}}
     & \multicolumn{2}{c}{\textbf{STS}}
     & \multicolumn{2}{c}{\textbf{KorSTS}} \\
    \cmidrule(lr){2-3}\cmidrule(lr){4-5}\cmidrule(lr){6-7}\cmidrule(lr){8-9}
    \textbf{Model}
     & before & after
     & before & after
     & before & after
     & before & after \\
    \midrule
    bge-en-icl
     & 0.1668 & \bfseries0.2574
     & \bfseries0.0511 & $-0.0745$
     & \bfseries0.8058 & 0.5965
     & \bfseries0.7078 & 0.2487 \\
    gte-Qwen2-1.5B-inst
     & \bfseries0.2858 & 0.2518
     & 0.0094 & \bfseries0.2204
     & \bfseries0.8592 & 0.7556
     & 0.3742 & \bfseries0.4727 \\
    e5-mistral-7b-inst
     & 0.1476 & \bfseries0.2641
     & \bfseries0.1099 & $-0.1738$
     & \bfseries0.8768 & 0.6092
     & \bfseries0.7495 & 0.1492 \\
    bge-large-en-v1.5
     & \bfseries0.1675 & 0.1626
     & $-0.2119$ & \bfseries$-0.1586$
     & 0.8752 & \bfseries0.8835
     & \bfseries0.3320 & 0.2473 \\
    all-MiniLM-L12-v2
     & 0.1909 & \bfseries0.2626
     & $-0.1837$ & \bfseries$-0.1590$
     & \bfseries0.8309 & 0.7109
     & \bfseries0.3858 & 0.1262 \\
    instructor-base
     & 0.2518 & \bfseries0.2646
     & $-0.0982$ & \bfseries$-0.0679$
     & \bfseries0.8585 & 0.8459
     & \bfseries0.0500 & 0.0116 \\
    bge-m3
     & 0.1969 & \bfseries0.2967
     & 0.0512 & \bfseries0.2732
     & \bfseries0.8194 & 0.7803
     & \bfseries0.7382 & 0.6919 \\
    \bottomrule
  \end{tabular}
\end{table*}

\begin{figure*}[!t]
\centering
\includegraphics[width=.93\textwidth]{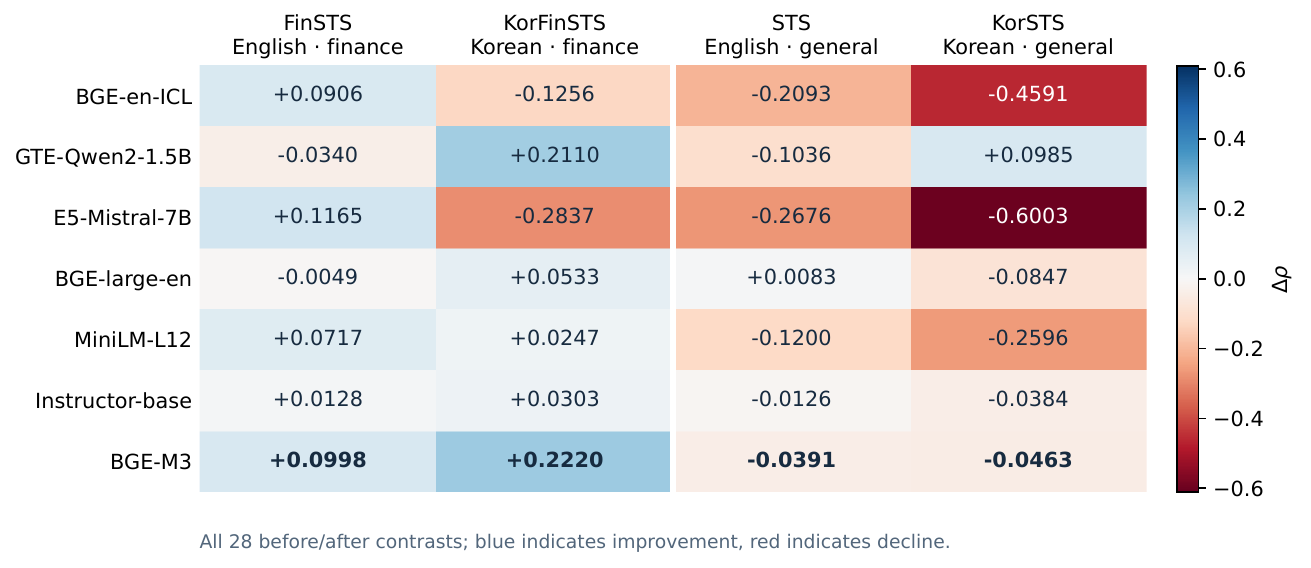}
\caption{Adaptation changes across the four evaluation settings. Financial gains vary across languages and encoders (RQ1), while general-domain losses are widespread (RQ2). Numerical labels show differences between the four-decimal scores in Table~\ref{tab:exp_results} on a common color scale.}
\Description{A seven by four heatmap labels each adaptation change. BGE-M3 improves both finance scores and loses both general scores.}
\label{fig:delta}
\end{figure*}

\begin{figure*}[!t]
\centering
\includegraphics[width=.94\textwidth]{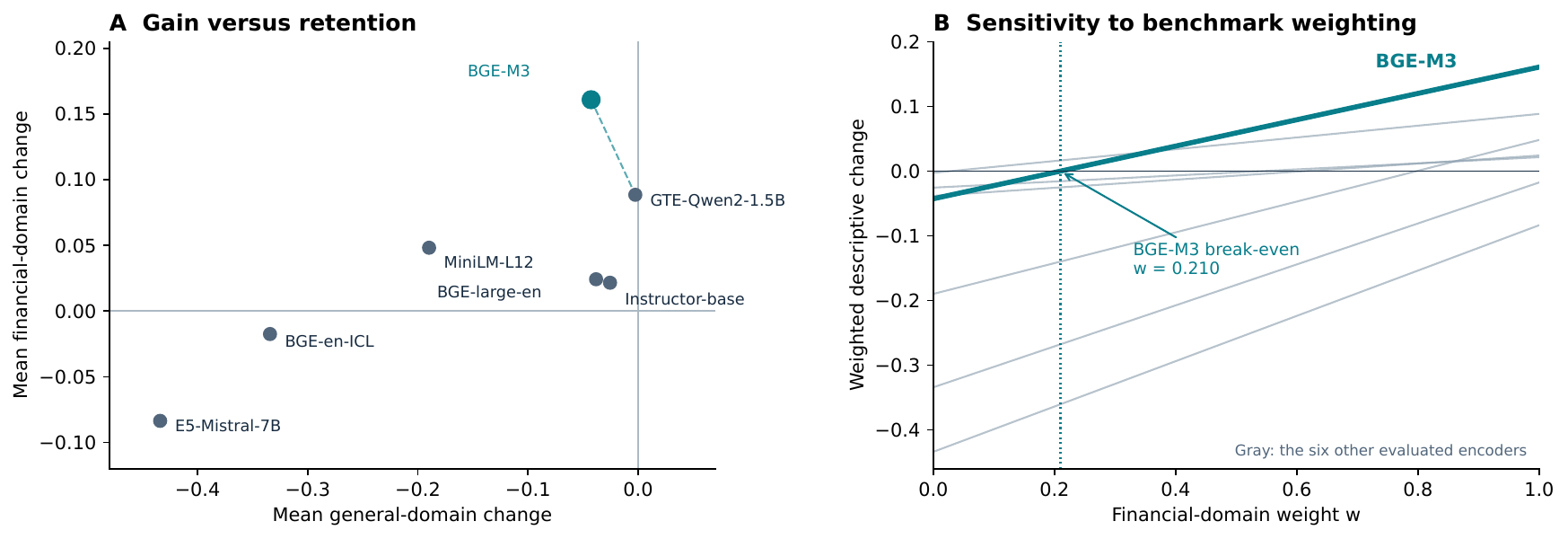}
\caption{Trade-off and sensitivity analysis of Table~\ref{tab:exp_results}. (A) Each point averages two financial and two general changes; the dashed segment connects the two nondominated adaptations. Higher and farther right is preferable under this summary. (B) A hypothetical weight $w$ interpolates between the general and financial means. This measures sensitivity to benchmark weights rather than deployment benefit.}
\Description{A scatter plot compares financial gain with general retention for seven models. A second plot varies the finance weight from zero to one. BGE-M3 crosses zero at about 0.210.}
\label{fig:tradeoff}
\end{figure*}

\paragraph{Training configuration.}
Each encoder is trained for one epoch on 18.8k triplets using four NVIDIA A100 GPUs, AdamW with learning rate $5\times10^{-5}$, and a linear warm-up over the first 10\% of steps. Batch sizes range from 8 to 64 according to memory requirements. The comparison applies a shared objective and data recipe; model size and memory requirements differ, so it does not hold compute cost constant.

\subsection{Evaluation Benchmarks}
We measure Spearman's $\rho$ on four suites: English financial FinSTS from FinMTEB~\citep{tang2025finmtebfinancemassivetext}; Korean financial KorFinSTS; general English STS from MTEB~\citep{muennighoff2023mtebmassivetextembedding}; and general Korean KorSTS from MMTEB~\citep{enevoldsen2025mmtebmassivemultilingualtext}.

\subsection{Results}
Table~\ref{tab:exp_results} presents the before/after correlations for all seven encoders. Figure~\ref{fig:delta} visualizes the 28 differences. We use $\Delta_{m,b}=\rho^{\mathrm{after}}_{m,b}-\rho^{\mathrm{before}}_{m,b}$ throughout. The four benchmark correlations remain separate measurements, allowing the two research questions to be examined by language.

\paragraph{RQ1: Financial gains depend on the backbone.}
Five of seven encoders improve on FinSTS and five improve on KorFinSTS, but only three improve on both: MiniLM-L12, Instructor-base, and BGE-M3. BGE-M3 has the strongest post-adaptation scores on the two financial suites, at 0.2967 and 0.2732. Its changes are $+0.0998$ and $+0.2220$, respectively. E5-Mistral improves more on FinSTS ($+0.1165$), illustrating the difference between the largest improvement and the highest final score. MiniLM and Instructor retain negative KorFinSTS correlations after adaptation despite positive changes; improvement alone is not evidence of adequate task performance.

\paragraph{RQ2: General capability is usually reduced.}
Six encoders decline on general English STS and six decline on KorSTS. The exceptions are BGE-large on English STS ($+0.0083$) and GTE-Qwen2 on KorSTS ($+0.0985$). Every encoder loses on the unweighted mean of the two general suites. BGE-M3 loses $0.0391$ and $0.0463$ on English and Korean general STS. BGE-en-ICL and E5-Mistral incur substantially larger losses, particularly on KorSTS ($-0.4591$ and $-0.6003$).

\section{Analysis and Practical Implications}
\subsection{Financial Gain and General Retention}
We summarize the benchmark-level changes using
\begin{align}
\Delta_F(m)&=\tfrac12(\Delta_{m,\mathrm{FinSTS}}+\Delta_{m,\mathrm{KorFinSTS}}),\\
\Delta_G(m)&=\tfrac12(\Delta_{m,\mathrm{STS}}+\Delta_{m,\mathrm{KorSTS}}).
\end{align}
These equal-weight summaries are not pooled correlations or deployment utilities. Per-language scores remain primary because averaging can obscure regressions.

Five models have positive $\Delta_F$: GTE-Qwen2 ($+0.0885$), BGE-large ($+0.0242$), MiniLM ($+0.0482$), Instructor ($+0.02155$), and BGE-M3 ($+0.1609$). BGE-en-ICL and E5-Mistral have negative financial means. BGE-M3 combines the highest financial mean gain with a general mean loss of $-0.0427$. GTE-Qwen2's nearly unchanged general mean ($-0.00255$) combines an English decline of $-0.1036$ with a Korean gain of $+0.0985$: cancellation across languages is not uniform retention.

Only GTE-Qwen2 and BGE-M3 are nondominated under these two change summaries: neither is exceeded on both summaries by another adaptation (Figure~\ref{fig:tradeoff}). Yet BGE-M3's final correlations exceed GTE-Qwen2's on all four suites. Change-space comparisons describe responses relative to different starting checkpoints; they do not identify equally capable final models. Both absolute scores and changes are therefore needed to interpret adaptation.

\subsection{Sensitivity to Domain Weighting}
For a transparent weighting check, define
\begin{equation}
U_m(w)=w\Delta_F(m)+(1-w)\Delta_G(m),\quad 0\leq w\leq1.
\end{equation}
Here $w$ is an analyst-selected emphasis on the two financial benchmarks, with equal English/Korean weights within each domain. For BGE-M3, $U(w)=0.2036w-0.0427$, which becomes nonnegative at $w\approx0.210$. At $w=0.5$, its four-suite mean change is $+0.0591$. The threshold describes sensitivity to benchmark priorities, not a deployment-benefit threshold: changing the language weights or tasks would change the summary.

\subsection{Implications for Financial Information Access}
Backbone variation cautions against predicting adaptation gains from language coverage alone. GTE-Qwen2 improves KorFinSTS by $0.2110$, while BGE-en-ICL and E5-Mistral regress on that benchmark despite improving FinSTS. Tokenization, pre-training, architecture, and optimization can all influence this response~\citep{wu2020languagescreatedequalmultilingual,wegmann2025tokenizationsensitivelanguagevariation}. Because these factors vary together across checkpoints, the comparison identifies adaptation outcomes rather than isolating a causal explanation for each backbone.

For an application considering specialization of an existing encoder, the results motivate three checks: improvement on the target financial language, acceptable absolute performance after adaptation, and retention on general-language tasks that share the encoder. A positive financial mean alone establishes none of these requirements. GTE-Qwen2 illustrates language-level cancellation, while MiniLM and Instructor illustrate the difference between an improvement and a satisfactory final score. Financial information-access systems should therefore retain per-language and general-domain evaluation alongside direct retrieval tests.

\section{Limitations}
\paragraph{Uncertainty and independence.}
The analysis uses aggregate correlations; prediction-level outputs and seed-level repeats are unavailable. We therefore report no confidence intervals or significance tests, and the 28 contrasts are not independent experimental replications. Exact temperature, random seeds, checkpoint versions, truncation settings, and per-model batch sizes are not fully documented, limiting exact reproducibility.

\paragraph{Dataset provenance and measurement.}
The source and length analysis characterizes the profiled dataset snapshot; its correspondence to the pairs underlying the aggregate correlations is not verified. This descriptive analysis does not validate label quality, annotation agreement, or source-level train--test independence. Synthetic variants and LLM judges may introduce shared errors, and aggregate results do not establish effectiveness per source or semantic axis.

\paragraph{Attribution and applicability.}
The experiment evaluates the combined recipe without ablations separating translation positives, paraphrases, negative types, or training budget. The seven backbones provide a bounded comparison rather than an exhaustive benchmark. Separate within-language STS scores do not establish cross-language retrieval, financial decision quality, or efficiency. These require task-specific evaluation.

\section{Conclusion}
NMIXX adapts existing encoders using source-linked financial meaning contrasts alongside paraphrase and Korean--English translation positives. Across seven backbones, the observed gains depend on both the model and the evaluation language. BGE-M3 achieves the highest adapted financial STS correlations in this comparison, but its gains coexist with general-domain losses. The principal implication is to evaluate financial adaptation through per-language gains, final capability, and retained general semantics together. The supervision design and KorFinSTS profile provide a basis for that evaluation; direct retrieval tests and controlled ablations are the next steps toward establishing application effectiveness and causal mechanisms.

\paragraph{Data availability.}
The \href{https://huggingface.co/datasets/nmixx-fin/NMIXX_train}{NMIXX training triplets} and \href{https://huggingface.co/datasets/nmixx-fin/korfinSTS}{KorFinSTS} are available on Hugging Face.

\section*{Acknowledgment}
Generative AI tools assisted manuscript organization, prose editing, and code for visualizations and descriptive dataset analysis. Their use in data construction is described in Section~3.

\bibliographystyle{ACM-Reference-Format}
\bibliography{software}
\end{document}